\documentclass[letterpaper, 10 pt, conference]{ieeeconf}  
\IEEEoverridecommandlockouts    
\usepackage{graphicx}
\usepackage[font=small]{caption}
\usepackage[font=small]{subcaption}
\usepackage{color}
\usepackage{algorithm}
\usepackage{algpseudocode}

\usepackage{siunitx}
\usepackage{textcomp}
\usepackage{gensymb}

\usepackage{amsfonts}
\usepackage{ dsfont }

\usepackage{amsmath}

\usepackage{soul}

\usepackage{url}

\usepackage{booktabs}

\usepackage{enumerate}  

\usepackage{fontawesome5}
\usepackage{tikz}

\usepackage{forest}

\usepackage[hyperfootnotes]{hyperref}  

\usepackage{textcase}
\usepackage[tablename=TABLE]{caption}
\DeclareCaptionTextFormat{up}{\MakeTextUppercase{#1}}
\usepackage{multirow}

\usepackage{lipsum} 

\title{ \large \bf  NTU4DRadLM: 4D Radar-centric Multi-Modal Dataset for Localization and Mapping}

\author{  \small
    Jun Zhang$^{*}$, Huayang Zhuge$^{*}$, Yiyao Liu$^{*}$, Guohao Peng$^{}$, Zhenyu Wu, Haoyuan Zhang, Qiyang Lyu,  Heshan Li, \\ Chunyang Zhao$^{}$,  Dogan Kircali, Sanat Mharolkar, Xun Yang, Su Yi, Yuanzhe Wang$^{+}$ and  Danwei Wang$^{}$    
	\thanks{All authors are with the School of Electrical and Electronic Engineering, Nanyang Technological University, Singapore. 
	    }%
        \thanks{* Co-first authorship, 
             {\tt\small \{jzhang061, hzhuge001, liuy0185\}@e.ntu.edu.sg}  }%
        \thanks{+ Corresponding author, {\tt\small \{yzwang, edwwang\}@ntu.edu.sg} } 
}

\begin{document}

\maketitle
\thispagestyle{empty}
\pagestyle{empty}

\begin{abstract}
Simultaneous Localization and Mapping (SLAM) is moving towards a robust perception age. However, LiDAR- and visual- SLAM  may easily fail in adverse conditions (rain, snow, smoke and fog, etc.). In comparison, SLAM based on 4D Radar, thermal camera and IMU can work robustly. But only a few literature can be found. A major reason is the lack of related datasets, which seriously hinders the research.  Even though some datasets are proposed based on 4D radar in past four years, they are mainly designed for object detection, rather than SLAM. Furthermore, they normally do not include thermal camera. Therefore, in this paper, \textit{NTU4DRadLM} is presented to meet this  requirement.
The main characteristics are: 1) It is the only dataset that simultaneously includes all $6$ sensors: 4D radar, thermal camera, IMU, 3D LiDAR, visual camera and RTK GPS. 
2) Specifically designed for SLAM tasks, which provides fine-tuned ground truth odometry and intentionally formulated loop closures.
3) Considered both low-speed robot platform and fast-speed unmanned vehicle platform.
4) Covered structured, unstructured and semi-structured environments. 
5) Considered both middle- and large- scale outdoor environments, i.e., the $6$ trajectories range from $246m$ to {$6.95km$}.
6) Comprehensively evaluated  three  types of SLAM algorithms.
Totally, the dataset is around {$17.6 km$, $85 mins$, $50 GB$} and it will be accessible from this link: \url{https://github.com/junzhang2016/NTU4DRadLM}

\end{abstract}


\section{Introduction}

\begin{figure}[t]
    \centering
    \begin{subfigure}{.45\textwidth}
		\centering
    	\includegraphics[width=1.0\linewidth]{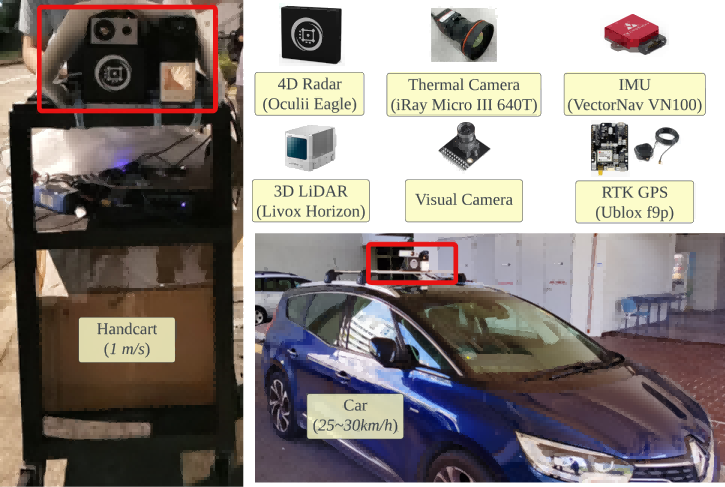}
    	\caption{The $6$ sensors and $2$ platforms}
    	\label{fig_intro_platform}
	\end{subfigure} 
    \begin{subfigure}{.26\textwidth}
		\centering
    	\includegraphics[width=1.0\linewidth]{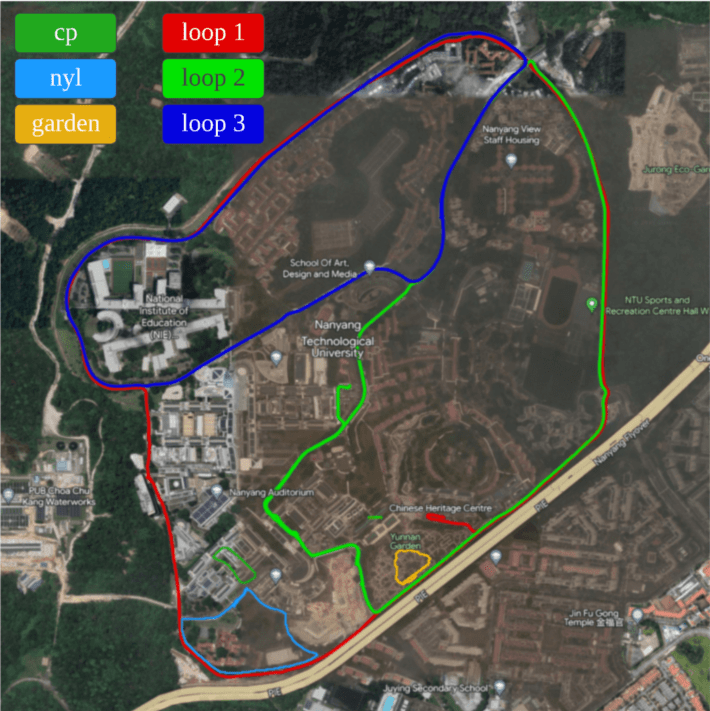}
    	\caption{ Six trajectories in NTU campus}
    	\label{fig_all_traj_google_earth}
	\end{subfigure}
    \begin{subfigure}{.185\textwidth}
        \centering
        \includegraphics[width=.98\linewidth]{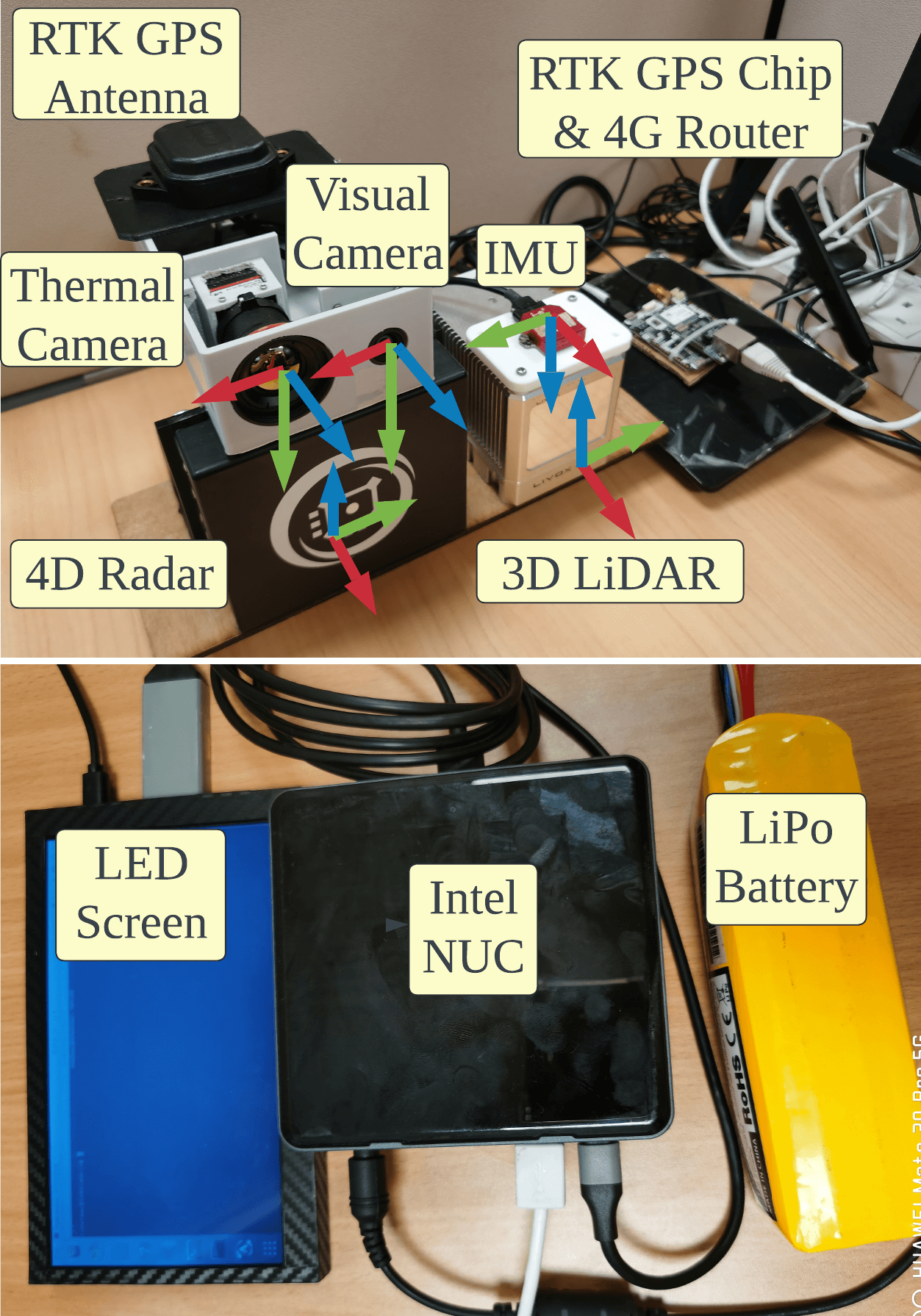}
        \caption{Sensor frames}
        \label{fig_sensor_frames_and_nuc}
    \end{subfigure} 
    \caption{A new dataset \textit{NTU4DRadLM} is presented to satisfy the urgent need of research on robust SLAM based on 4D radar, thermal camera and IMU. (a) The $6$ sensors and the slow- and fast-  speed platforms: a handcart ($1m/s$) and a car ($25-30km/h$). (b) Satellite image of the six trajectories plotted on Google map, three are collected with the handcart, another three are with the car. (c) The sensor frames, red: $x-$axis, green: $y-$axis, blue: $z-$axis.   Mini-computer, LED screen and battery.}  %
    \label{fig_intro}
    \vspace{-16pt}
\end{figure}

Simultaneous Localization and Mapping (SLAM) is one of the building blocks of autonomous mobile robots~\cite{KongHe2021ResourceAR, KongHe2021TRO, feng2023exploring} and unmanned vehicles~\cite{apollo2017,autoware201X}.
Currently, most research is focused on LiDAR- and visual- SLAM~\cite{Qin2018TRO,Shan2020IROS}. However, these sensors may not cope well with adverse conditions (e.g., heavy rain, snow, smoke, fog and dust, etc.). Therefore, robust SLAM in adverse conditions becomes more and more important.



Fortunately, a new sensor comes into the market - \textbf{4D imaging radar}. It can output dense 3D point cloud with elevation information. By combining it with \textbf{thermal camera} and \textbf{IMU}, which are also robust to adverse conditions, robust SLAM in adverse conditions can be achieved. However, only a few related  works can be found~\cite{Ng2021IROS,Doer2021IROS,Zhang2023ICRA,Zhang2023CISRAM,Zhuang2023RAL}. One main reason is the lack of related datasets, which simultaneously include 4D radar, thermal camera, IMU and ground truth odometry. This is not surprising, since: 1) 4D Radar is a relatively new sensor, and not cheap for now. 2) Thermal camera is normally  more expensive than visual camera, and it is not easy to extract enough features from thermal images. The lack of datasets seriously hinders related research. Thus, we propose this dataset  to promote SLAM research based on 4D radar, thermal camera and IMU. 





Compared with existing 4D radar datasets, the uniqueness and contributions of ours can be found in Tab.\ref{table_datasets_comparison} and briefly described below: 


\begin{enumerate}    
    \item \textit{This is the only dataset that simultaneously includes all $6$ sensors and the calibration parameters.} Apart from ours, only RRxIO~\cite{Doer2021IROS} includes a thermal camera, but it is mainly proposed for Unmanned Aerial Vehicle (UAV), small-scale environments and no 3D LiDAR and GPS included. 
    \item \textit{Specifically designed for SLAM tasks.}  We not only provide fine-tuned ground truth odometry, but also intentionally traversed partly overlapped trajectory to formulate loop closures for graph optimization. In comparison, most existing datasets are oriented for object detection~\cite{apalffy2022RAL},~\cite{Zheng2022ITSC}, and  do not consider loop closures. 
    \item \textit{Considered both low-speed robot platform and fast-speed unmanned vehicle platform.} In comparison,  most datasets only use one platform, either low-speed robot~\cite{Doer2021IROS},~\cite{Kramer2022IJRR}  or fast-speed vehicle~\cite{Rebut2022CVPR},~\cite{apalffy2022RAL},~\cite{Zheng2022ITSC},~\cite{paek2022kradarNIPS}. 
    \item \textit{Covered structured, unstructured and semi-structured environments,} i.e., structured carpark near academic buildings, unstructured garden, and semi-structured campus main road. In comparison,  most existing datasets only cover {either one type~\cite{Meyer2019EuRAD,Doer2021IROS,Rebut2022CVPR,apalffy2022RAL,Zheng2022ITSC,paek2022kradarNIPS} }.
    \item \textit{Considered middle-scale and large-scale outdoor environments,} i.e., the $6$ trajectories range from $246m$ to {$6.95km$}.
    \item \textit{Comprehensive evaluation of three types of 4D Radar SLAM,} i.e., pure 4D Radar SLAM, 4D Radar-IMU SLAM, and 4D Radar-thermal camera SLAM, which has not been compared in existing datasets.
\end{enumerate}

\begin{table*}[t]
    \centering
    \caption{Comparison of existing 4D Radar datasets and our dataset. \textbf{3DL}: 3D LiDAR, \textbf{VC}: Visual Camera,   \textbf{4DR}: 4D RADAR, \textbf{TC}: Thermal Camera, \textbf{SLAM}: Simultaneous Localization and Mapping,  \textbf{OD}: Object Detection, \textbf{Struc.}: Structured, \textbf{Unstruc.}: Unstructured. } 
    \begin{tabular}{c|cccccc|cc|cc|ccc}
    \multirow{2}{*}{Dataset} & \multicolumn{6}{c|}{Sensors} & \multicolumn{2}{c|}{Task Oriented} & \multicolumn{2}{c|}{Platform Speed}     & \multicolumn{3}{c}{Environments} \\ \cline{2-14} 
     & 3DL  & VC  & 4DR  & TC & IMU & GPS & \multicolumn{1}{c}{SLAM} &  OD & \begin{tabular}[c]{@{}c@{}}Robot \\ (low)\end{tabular} & \begin{tabular}[c]{@{}c@{}}Vehicle\\ (high)\end{tabular} & Scale    & Struc.   & Unstruc.   \\ \hline
    Astyx~\cite{Meyer2019EuRAD}    &  \checkmark  &   \checkmark  &    \checkmark  &   $\times$  &  $\times$  &  $\times$  &   $\times$  &  \checkmark  &  $\times$  &  \checkmark  &  small  &   \checkmark    &  $\times$  \\
    RRxIO~\cite{Doer2021IROS}    & $\times$   &  \checkmark   &   \checkmark &  \checkmark   &  \checkmark &  $\times$    &  \checkmark  &  $\times$  &  \checkmark  &  $\times$  &  small  &    \checkmark   &  $\times$  \\
    RADIal~\cite{Rebut2022CVPR}     &  \checkmark  &   \checkmark  &   \checkmark  &  $\times$   &  $\times$ &  \checkmark  &  \checkmark  & \checkmark  &  $\times$  &  \checkmark  &  middle  &   \checkmark    &  $\times$  \\
    ColoRadar~\cite{Kramer2022IJRR}    &  \checkmark  &  $\times$   &   \checkmark  &  $\times$   &  \checkmark  &  $\times$ & \checkmark   &   $\times$  &  \checkmark  &  $\times$   &  middle, large &  \checkmark     &  \checkmark  \\
    VoD~\cite{apalffy2022RAL}    &  \checkmark  &  \checkmark   &  \checkmark &  $\times$  & \checkmark &  \checkmark &  \checkmark  & \checkmark   &  $\times$   &  \checkmark  &  middle  &   \checkmark    & $\times$    \\
    TJ4DRadSet~\cite{Zheng2022ITSC}     &  \checkmark  &  \checkmark   &  \checkmark   & $\times$     & $\times$  & \checkmark &  \checkmark  &  \checkmark  & $\times$    &  \checkmark  & middle   &    \checkmark   & $\times$    \\
    K-Radar~\cite{paek2022kradarNIPS}    &  \checkmark  & \checkmark    &  \checkmark  &  $\times$    &  \checkmark  & \checkmark  &  \checkmark   &   \checkmark  & $\times$    & \checkmark   & middle   &     \checkmark  &  $\times$  \\ \hline
    \textbf{Ours (NTU4DRadLM)}    &  \checkmark  & \checkmark     & \checkmark   &  \checkmark   & \checkmark & \checkmark  & \checkmark   &  $\times$    & \checkmark   & \checkmark   & middle, large   & \checkmark       & \checkmark    
    \end{tabular}
    \label{table_datasets_comparison}
\end{table*}

The paper is organized as follows: Section \ref{sec_methodology} introduces the details of the proposed dataset. Section \ref{sec_experimental_results} demonstrates experiments and analysis of three types SLAM methods. Finally, Section \ref{sec_conclusions_and_future_work} concludes the paper and discusses future work.

\begin{table*}[t]
    \centering
    \caption{The sensors used and the specifications. }
    \begin{tabular}{c|ccccccc}
    \hline
    Sensor       & Type    & Description     & Data & Hz   &   {\begin{tabular}[c]{@{}c@{}}Range\\ and error\end{tabular}}  &   {\begin{tabular}[c]{@{}c@{}}FOV\\ ($H\times V$)\end{tabular}}    &     {\begin{tabular}[c]{@{}c@{}}Resolution\\ ($H\times V$)\end{tabular}} \\ \hline
    3D LiDAR     & Livox Horizon     & Non-repetitive scanning & Ethernet   & 10  &    {\begin{tabular}[c]{@{}c@{}}$260m$\\ $(\pm 
 2cm)$\end{tabular}}   & $81.7 \degree \times 25.1 \degree $  & time-varying    \\
    Visual Camera  & Vishinsgae SY020HD     & Web camera    & USB2.0   & 30  & -     &  $88 \degree \times 66 \degree $    & $640 \times 480$ (pixel)   \\
    4D Radar     & Oculii Eagle      & \{x,y,z,doppler,power\} & Ethernet   & 12  & {\begin{tabular}[c]{@{}c@{}}$400m$\\ $(\leq 0.86m)$ \end{tabular}}   & $120 \degree \times 30 \degree $  & $0.5 \degree \times 1 \degree$    \\
    Thermal Camera & iRay Micro III 640T & Uncooled infrared    & USB3.0   & 25  & -     & $70 \degree \times 57 \degree $     & $640 \times 512$ (pixel)    \\
    IMU    & Vector VN-100     & 9-axis    & USB2.0   & 200 & -     & -     & -    \\
    RTK GPS      & Ublox F9P-02B-00    & SiReNT RTK Subscription     & USB2.0   & 20    & -     & -     & -    \\ \hline
    \end{tabular}    
    \label{tab_sensors_description}
\end{table*}

\section{The NTU4DRadLM Dataset} \label{sec_methodology}


\subsection{ Sensors and Platform }
\subsubsection{Sensors}
The sensor suite and sensor frames are depicted in Fig.\ref{fig_intro_platform} and Fig.\ref{fig_sensor_frames_and_nuc}. It consists of $6$ heterogeneous sensors: a 3D LiDAR, a visual camera, a 4D Radar, a thermal camera, an IMU and a RTK GPS. The specifications of the sensors are shown in Tab.\ref{tab_sensors_description}.

\subsubsection{Platform}
To satisfy the requirement of both low-speed mobile robot and fast-speed unmanned vehicle, we collect the dataset with two platforms, as shown in Fig.\ref{fig_intro_platform}: a handcart that moves at the speed of most mobile robots (about $1m/s$) and a multi-purpose vehicle (MPV) (about $25-30km/h$). 



A mini-computer is connected to all sensors to collect data. All devices are powered by LiPo battery, as shown in Fig.\ref{fig_sensor_frames_and_nuc}. The mini-computer is Intel® NUC NUC10i7FNH, with {32GB RAM, 1TB SSD}, Ubuntu 18.04 and ROS melodic. 


\begin{figure}[h]
    \centering
    \begin{subfigure}{.40\textwidth}
		\centering
		\includegraphics[width=0.99\linewidth]{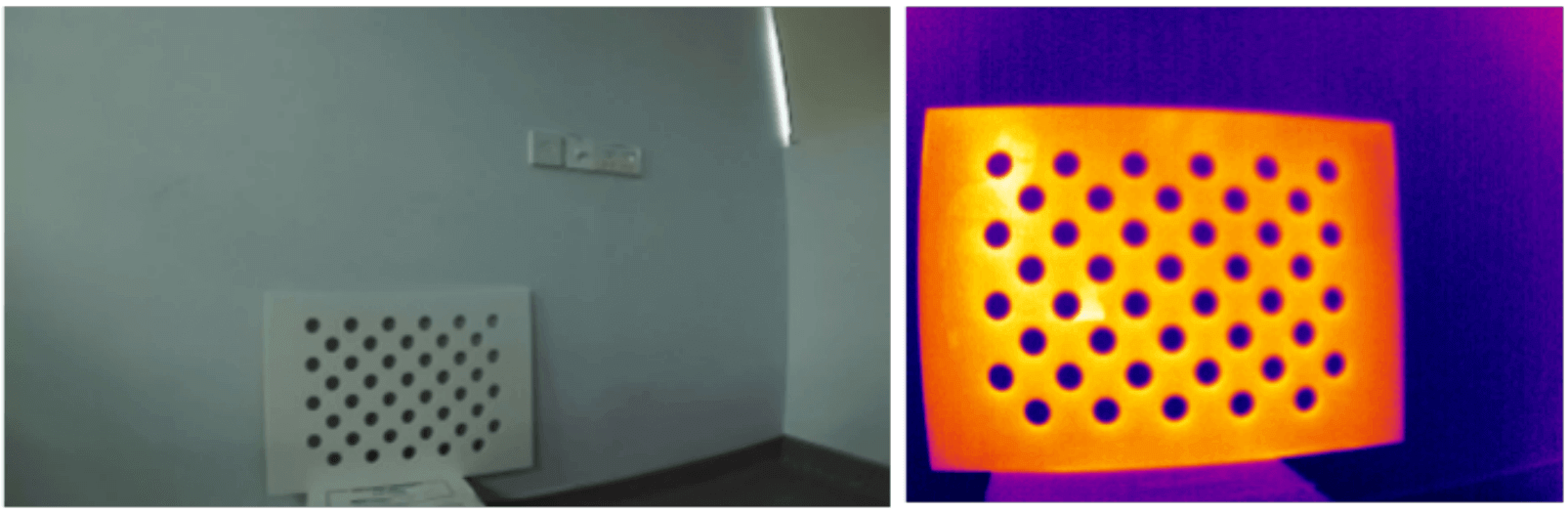}
        \caption{4x11 asymmetric circular holes board~\cite{Zhang2018ICARCV}.}
		\label{fig_4x11_circles_board}
	\end{subfigure} \\
	\begin{subfigure}{.20\textwidth}
		\centering
		\includegraphics[width=0.95\linewidth]{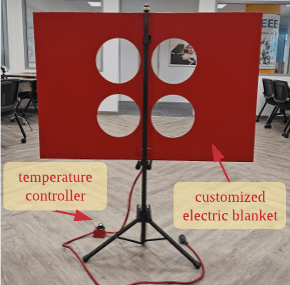}
        \caption{Four-circular-hole board~\cite{Zhang2023IVS}.}
		\label{fig_four_circle_holes}
	\end{subfigure}
    \begin{subfigure}{.18\textwidth}
		\centering
		\includegraphics[width=0.99\linewidth]{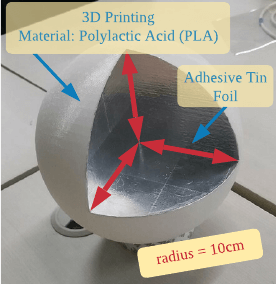}
		\caption{Spherical-trihedral~\cite{Zhang2022ITSC}.}
		\label{fig_spherical_trihedral}
	\end{subfigure}
	
    \caption{ The targets used for intrinsic and extrinsic calibration.  }
    \label{fig_calib_target}
    \vspace{-16pt}
\end{figure}

\subsection{Calibration}
\subsubsection{Intrinsic   Calibration of  Visual and Thermal Camera}

As shown in Fig.\ref{fig_4x11_circles_board}, a rectangular board with $4\times11$  asymmetric circular holes  is employed to  obtain the intrinsic  parameter of  visual and thermal camera.  The process is based on the well-known Zhang's method~\cite{Zhang's2000}. More details can be referred to our previous work~\cite{Zhang2018ICARCV}. 



\subsubsection{Intrinsic Calibration of IMU}
The ROS package \texttt{IMU\_utils}~\cite{IMUutils2018git} is adopted.  The IMU comprises an accelerator and a gyroscope. The white noise and bias of both accelerator and gyroscope can be obtained.



\subsubsection{Extrinsic Calibration of LiDAR-Thermal-Visual}
As shown in Fig.\ref{fig_four_circle_holes}, a rectangular board with four-circular-holes is utilized for the extrinsic calibration between LiDAR-thermal and LiDAR-visual camera. We open-source the code on github as \texttt{lvt2calib}~\cite{lvt2calib2023git}. More details  can be referred to our paper~\cite{Zhang2019ROBIO,Zhang2023IVS}.


\subsubsection{Extrinsic Calibration of 4D Radar-Thermal}
As shown in Fig.\ref{fig_spherical_trihedral}, a spherical-trihedral target is used for the extrinsic calibration. The target is heated up by a hair-dryer for thermal camera to detect. The sphere center is extracted as the common feature. By minimizing 2D-3D re-projection error, optimal extrinsic parameter can be obtained. More details can be found in  our previous work~\cite{Zhang2022ITSC}.




\subsubsection{Temporal and Extrinsic Calibration of LiDAR-IMU}
\texttt{LiDAR\_IMU\_Init}~\cite{LiDARIMUInit} is adopted. It is designed for Livox-type LiDAR, thus a perfect fit to our configuration. It can simultaneously output both temporal offset and extrinsic parameter. It is also very convenient to use since it can  automatically detect the degree of excitation and instruct users to give sufficient excitation.


\subsubsection{Extrinsic Calibration of GPS-Init}

\textcolor{black}{In order to make use of GPS coordination, extrinsic calibration between GPS UTM coordinate and the ROS xyz coordinate is required. Current works use the first  frame point cloud as origin in default. Init is the origin that set the first LiDAR frame pose as 0. We design a graph optimization method to find the transformation matrix that transforms GPS UTM coordinate to Init coordinate.} Firstly, ground truth odometry and GPS message are associated based on timestamps. Then, convert GPS message (latitude, longitude, altitude) into UTM coordinate (easting, northing, upward). Construct a pose graph with a single node representing the transformation matrix, and $SE(3)$ edges transforming all UTM coordinates to corresponding ground truth positions. Finally, optimize the graph to get the optimal estimation.

\subsubsection{Calibration Evaluation}
\begin{figure}[!h]
	\centering
	\begin{subfigure}{.2\textwidth}
		\centering
    	\includegraphics[width=1.0\linewidth]{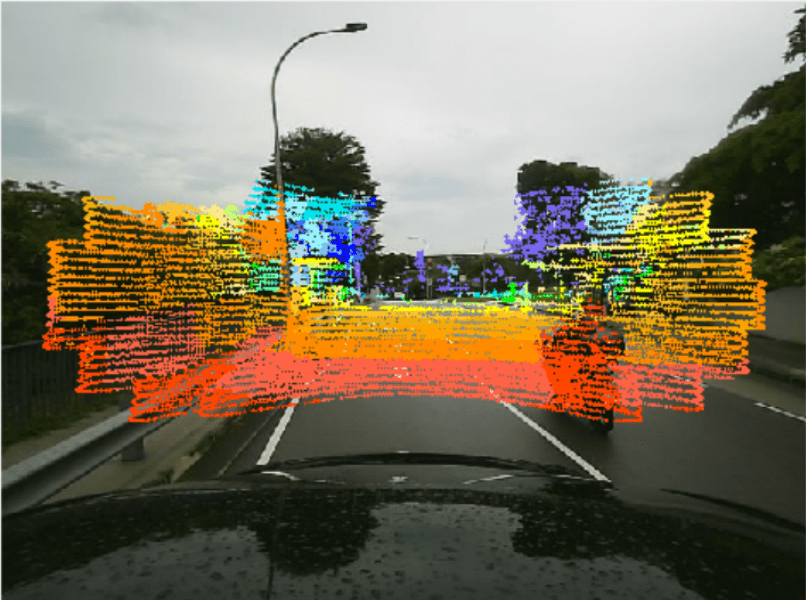}
    	\caption{}
    	\label{fig_projection_lidar_rgb}
	\end{subfigure} 
	\begin{subfigure}{.2\textwidth}
		\centering
    	\includegraphics[width=0.98\linewidth]{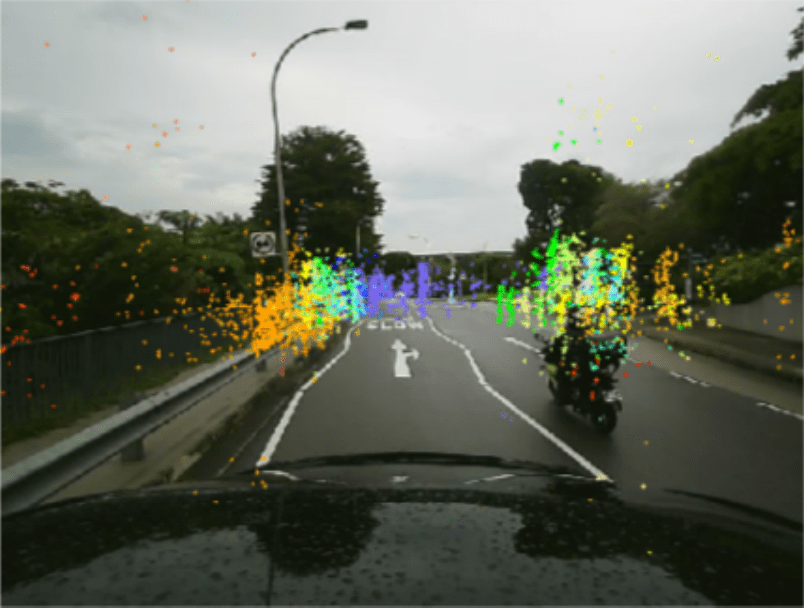}
    	\caption{}
    	\label{fig_projection_radar_rgb}
	\end{subfigure}  \\
      \begin{subfigure}{.2\textwidth}
		\centering
    	\includegraphics[width=1.0\linewidth]{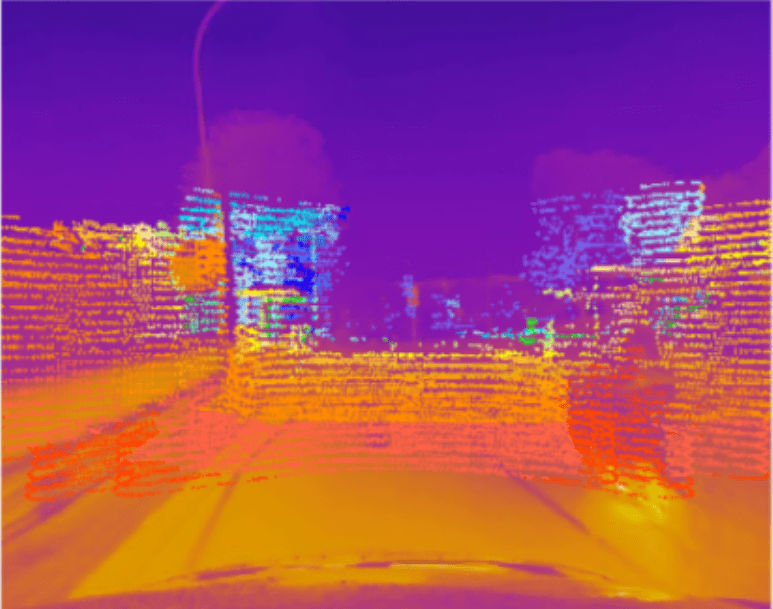}
    	\caption{}
    	\label{fig_projection_lidar_thermal}
	\end{subfigure}
	\begin{subfigure}{.2\textwidth}
		\centering
    	\includegraphics[width=0.98\linewidth]{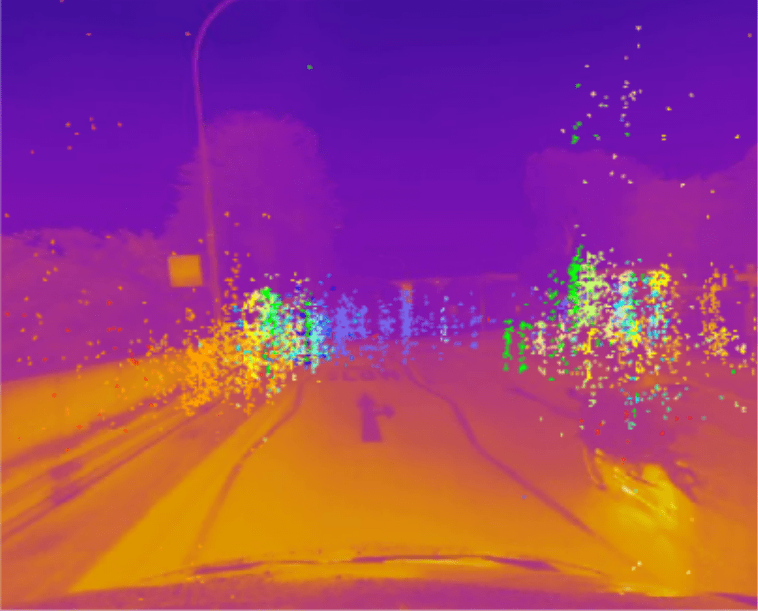}
    	\caption{}
    	\label{fig_projection_radar_thermal}
	\end{subfigure} \\
      \begin{subfigure}{.4\textwidth}
		\centering
    	\includegraphics[width=1.0\linewidth]{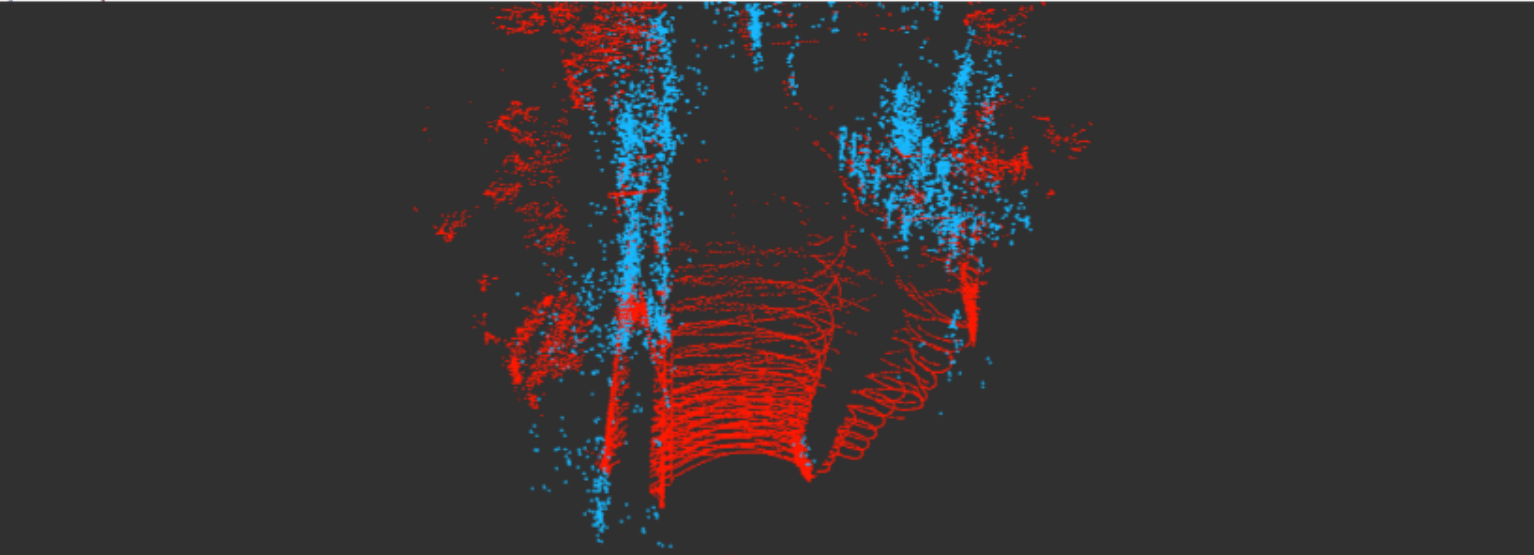}
    	\caption{ }
    	\label{fig_projection_lidar_radar}
	\end{subfigure}
	\caption{Projection and registration results after calibration. (a) Lidar-rgb. (b) Radar-rgb. (c) Lidar-thermal. (d) Radar-thermal. (e) Lidar-Radar  (red: LiDAR, blue: Radar).}
	\label{fig_projection}
	\vspace{-12pt}
\end{figure}

With the intrinsic and extrinsic parameters, the LiDAR and Radar point cloud can be projected onto RGB and thermal image, respectively, as shown in Fig.\ref{fig_projection_lidar_rgb}, Fig.\ref{fig_projection_radar_rgb}, Fig.\ref{fig_projection_lidar_thermal} and Fig.\ref{fig_projection_radar_thermal}. The LiDAR and Radar point cloud can be transformed together, as shown in Fig.\ref{fig_projection_lidar_radar}.


\subsection{Data Collection} 
\begin{figure}[!t]
	\centering
	\begin{subfigure}{.17\textwidth}
		\centering
    	\includegraphics[width=1.0\linewidth]{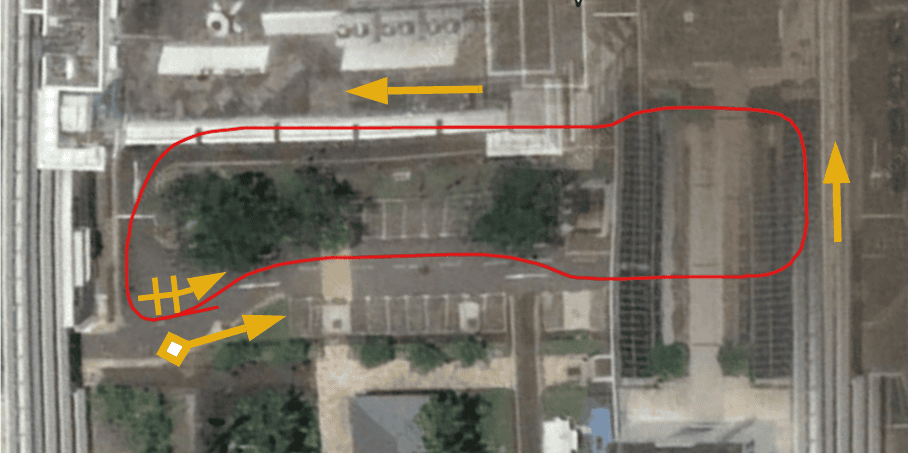}
    	\caption{  cp ($246m$)}
    	\label{fig_cp_google_earth}
	\end{subfigure} 
	\begin{subfigure}{.16\textwidth}
		\centering
    	\includegraphics[width=1.0\linewidth]{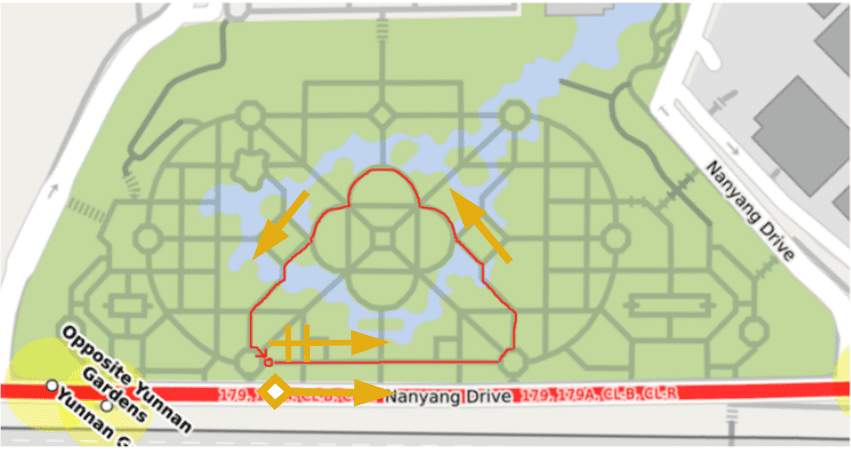}
    	\caption{   garden ($339m$)}
    	\label{fig_garden_google_earth}
	\end{subfigure} 
      \begin{subfigure}{.14\textwidth}
		\centering
    	\includegraphics[width=1.0\linewidth]{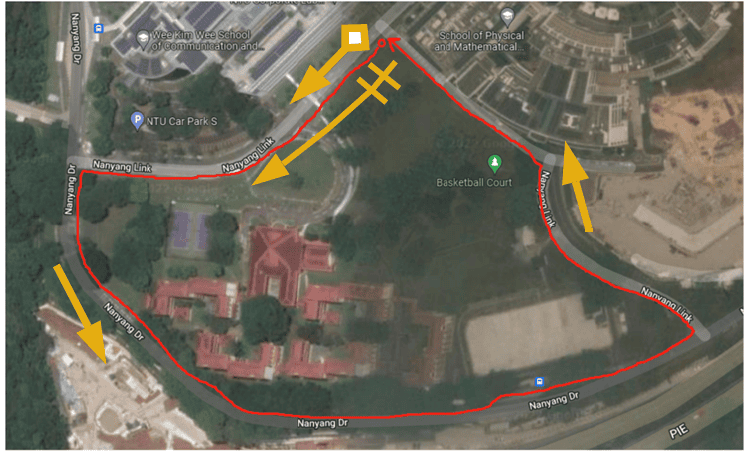}
    	\caption{ nyl ($1017m$)}
    	\label{fig_nyl_google_earth}
	\end{subfigure} \\
     \begin{subfigure}{.15\textwidth}
		\centering
    	\includegraphics[width=.98\linewidth]{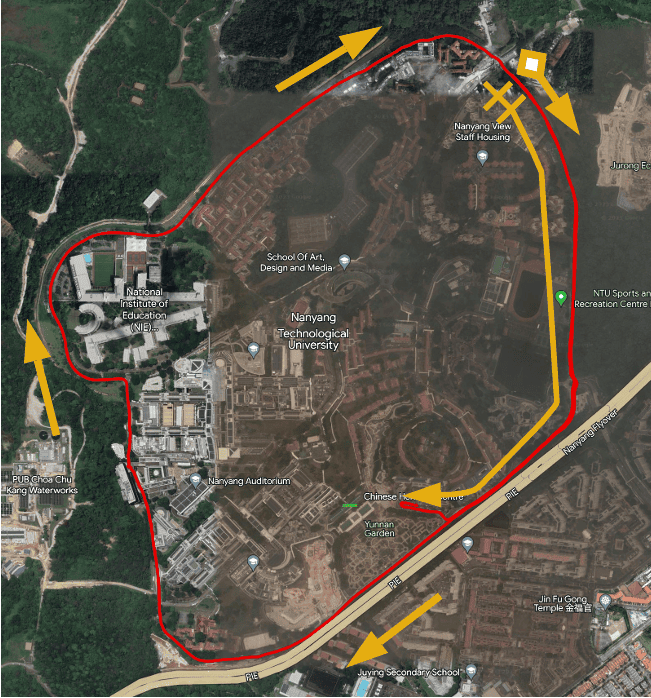}
    	\caption{ loop 1 ($6.95km$)}
    	\label{fig_loop1_google_earth}
	\end{subfigure}
	\begin{subfigure}{.16\textwidth}
		\centering
    	\includegraphics[width=1.0\linewidth]{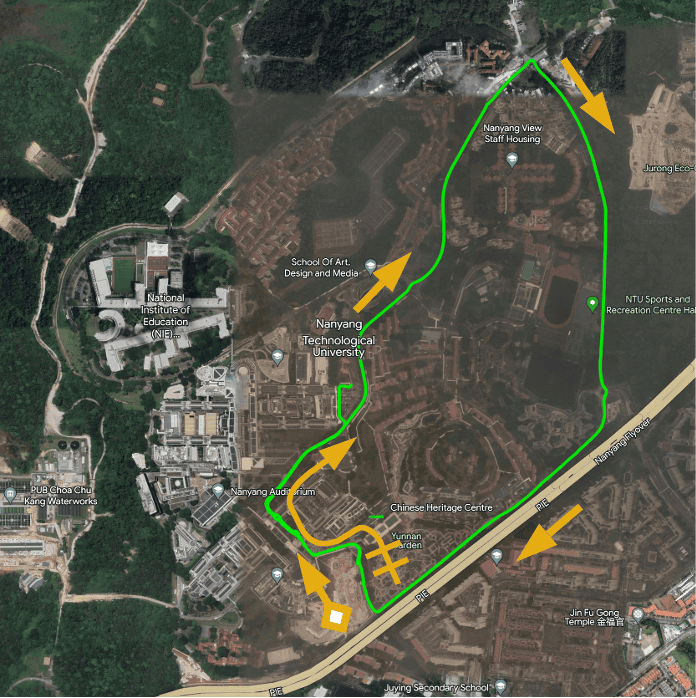}
    	\caption{  loop 2 ($4.79km$)}
    	\label{fig_loop2_google_earth}
	\end{subfigure} 
      \begin{subfigure}{.16\textwidth}
		\centering
    	\includegraphics[width=.97\linewidth]{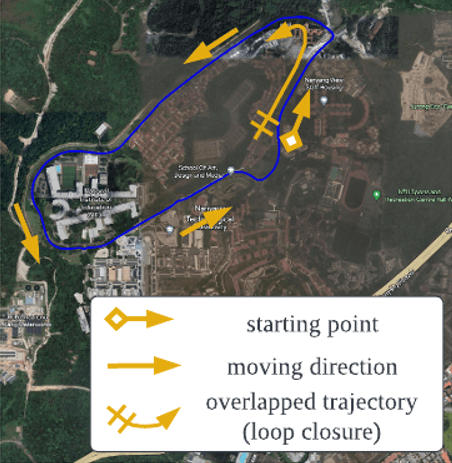}
    	\caption{ loop 3 ($4.23km$)}
    	\label{fig_loop3_google_earth}
	\end{subfigure}
     
	\caption{The satellite image of the $6$ trajectories we traversed in the NTU campus. The starting point, moving direction and overlapped trajectory part are shown in orange arrows. (a)(b)(c) are collected with the handcart. (d)(e)(f) are collected with the car. }
	\label{fig_trajectory}
	\vspace{-16pt}
\end{figure}

As mentioned before, we used two platforms to collect the data: a handcart  and a car. This allowed us to collect data in both small- and large- scale environments, as well as structured and unstructured environments, with low- and fast- speeds. A summary of the $6$ datasets is shown in Tab.~\ref{tab_our_datasets_summary}. The satellite image of the $6$ trajectories are separately presented in  Fig.~\ref{fig_trajectory} and plotted together in Fig.~\ref{fig_all_traj_google_earth}. 



\subsubsection{Handcart Platform}
For the handcart platform, we collect dataset in three scenarios: NTU Carpark P (\texttt{cp}), Yunan Garden (\texttt{garden}), and Nanyang Link (\texttt{nyl}). The satellite image of the three routes can be found in Fig.\ref{fig_cp_google_earth}, Fig.\ref{fig_garden_google_earth}, Fig.\ref{fig_nyl_google_earth}. The routes cover structured, unstructured, and semi-structured environments, respectively. The trajectory length is $246m$, $339m$ and $1017m$. The handcart is pushed by human. The average moving speed is around $1m/s$. 

Some raw data samples collected with the handcart can be found in column $1-3$ of Fig.\ref{fig_data_sample}.

\subsubsection{Car Platform}
For the car platform, we collect dataset in three routes in the campus main road: \texttt{loop 1}, \texttt{loop 2}, and \texttt{loop 3}. The satellite image of the three routes can be found in Fig.\ref{fig_loop1_google_earth}, Fig.\ref{fig_loop2_google_earth}, Fig.\ref{fig_loop3_google_earth}. The trajectory length is {$6.95km$}, $4.79km$, and $4.23km$. The vehicle is driven smoothly by human driver, with an average speed of about $25-30km/h$. 

\begin{figure*}[htbp]
    \centering
    \includegraphics[width=0.9\linewidth]{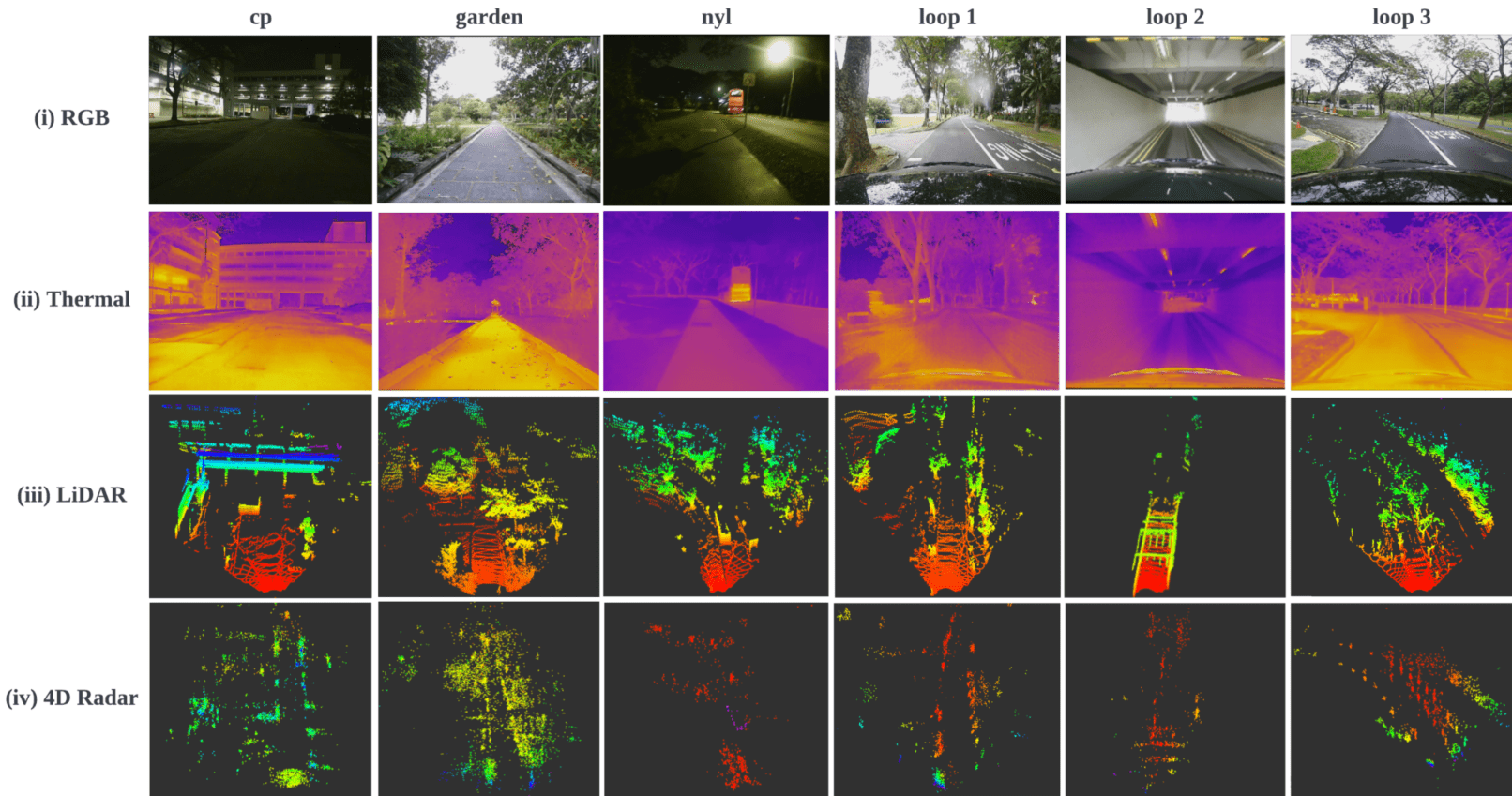}
    \caption{Visualization of the raw data samples of the six trajectories.}  
    \label{fig_data_sample}
\end{figure*}

Some raw data samples collected with the car can be found in column $4-6$ of Fig.\ref{fig_data_sample}.

While collecting the datasets, we take some precautions:
\begin{itemize}
    \item Before moving, we keep the whole platform static for about $5$ seconds, we move the platform as smooth as possible to avoid too sudden start and stop, as well as too sharp turns. 
    \item We also considered loop closures, which is important for graph optimization to correct the odometry drift error. We intentionally traverse overlapped trajectories to formulate more loop closures, i.e., continue moving forward after returning to the starting position.
    \item To ensure the data collection reliability,  we set the rosbag to be automatically splitted once a rosbag reaches $3GB$. We also set a buffer size of $3GB$ in case of data loss.  The  command is: \texttt{rosbag record -b 3072 --split --size 3072 }.
\end{itemize}

     

\begin{table}[!hb]
    \centering
    \caption{Summary of the $6$ datasets, \textbf{Stru/Unstruc}: structured or unstructured environment.   } 
     \label{tab_our_datasets_summary}
     \resizebox{0.45\textwidth}{!}{
    \begin{tabular}{ccllll}
    \hline
    \multicolumn{1}{l}{Platform} & \multicolumn{1}{c}{Speed}    & Name      &  {\begin{tabular}[c]{@{}c@{}}Length (Duration)\end{tabular}}  & GPS & Stru/Unstruc \\ \hline
    \multirow{4}{*}{handcart}     & \multirow{4}{*}{$\approx$1 m/s} & cp    &  {\begin{tabular}[c]{@{}c@{}}246 m  (7m:16s)\end{tabular}}    & no    & struc.      \\
     &    & garden    & {\begin{tabular}[c]{@{}c@{}}339 m  (11m:27s)\end{tabular}}   & no    & unstruc.    \\
     &    & nyl       & {\begin{tabular}[c]{@{}c@{}}1017 m   (20m:03s)\end{tabular}}  & no    & semi-struc.   \\ \hline
    \multirow{5}{*}{car}       & \multicolumn{1}{l}{30 km/h}  & loop 1    &   {\begin{tabular}[c]{@{}c@{}}6.95 km   (22m:51s)\end{tabular}}  & yes    & semi-struc.  \\
    & \multicolumn{1}{l}{25 km/h}  & loop 2    &  {\begin{tabular}[c]{@{}c@{}}4.79 km  (16m:49s)\end{tabular}}  & yes    & semi-struc.   \\
     & \multicolumn{1}{l}{30 km/h}  & loop 3    & {\begin{tabular}[c]{@{}c@{}}4.23 km   (10m:44s)\end{tabular}}   & yes    & semi-struc.  \\
     
    \hline
    \end{tabular}
    }
\end{table}


\subsection{Ground Truth Odometry}

The ground truth trajectory is obtained by a tightly-coupled LiDAR-Visual-Inertial SLAM \texttt{R2LIVE}~\cite{Lin2021RAL_r2live}.  However, we found there exists trajectory drift for large scale environments. To solve the problem, we construct a pose graph optimization to correct the drift. Loop closures  are formed with several pairs of overlapping points on the trajectory. \texttt{g2o}~\cite{g2o2011ICRA} is used to calculate the optimal results.

\subsection{Data Structure}
\begin{figure}[!htb]
    \centering
    \includegraphics[width=0.9\linewidth]{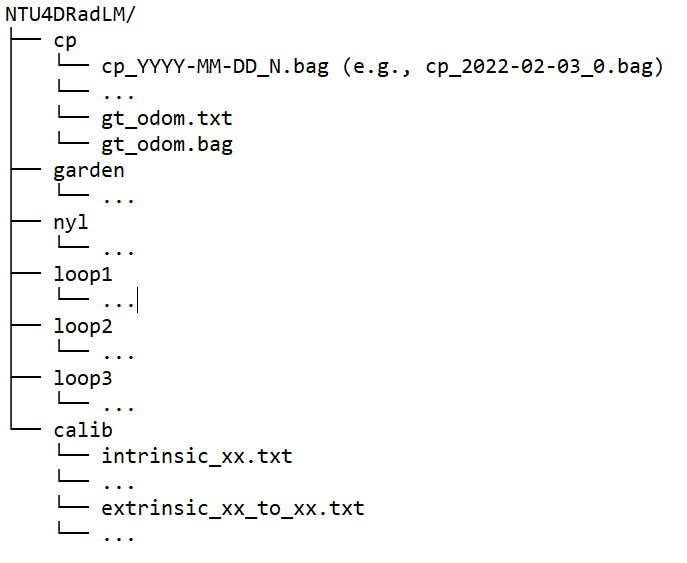}
    \caption{The data structure.}
    \label{fig_data_structure}
\end{figure}

The data structure of the collected datasets is depicted in Fig.\ref{fig_data_structure}. Under the folder NTU4DRadLM, there are $7$ folders: six folders to store the rosbags and ground truth odometry for the six routes, one folder to store the calibration parameters (both intrinsic and extrinsic). 

Taking the route \texttt{cp} as an example: all raw data is saved as ROS topics in rosbag. Each rosbag is named in the format ``ROUTE\_NAME\_YYYY-MM-DD\_N". The ``ROUTE\_NAME" denotes the route name (e.g., cp, garden, nyl, ...), followed by the date ``YYYY-MM-DD", and the last digit ``N" denotes the $N_{th}$ rosbag (e.g., 0, 1, 2, ...). The ground truth odometry is saved as ``gt\_odom.txt" and ``gt\_odom.bag", the former is generated from the latter. We use the \texttt{rpg\_trajectory\_evaluation} ~\cite{rpgtrajeval2018,Zhang18IROS_rpg}.




As for the ``calib" folder, it saves both intrinsic and extrinsic parameters. ``intrinsic\_xx.txt" denotes the intrinsic parameter, ``xx" can be: RGB camera, thermal camera, and IMU. ``extrinsic\_xx\_to\_xx.txt", denotes the extrinsic from one sensor to another sensor. We follow the KITTI format~\cite{Geiger2013IJRR}.

\section{ Evaluation of NTU4DRadLM Dataset} \label{sec_experimental_results}

\begin{figure}[b]
    \centering
     \resizebox{0.45\textwidth}{!}{
    \begin{forest}
  [ \small \textbf{Three Types of SLAM} 
    [\small Pure 4D Radar 
      [\small gicp] 
      [\small gicp+{lc}] 
    ]
    [\small 4D Radar-IMU  
      [\small fast\_lio] 
    ]
    [\small 4D Radar-Thermal 
      [\small 4DRT\_SLAM] 
    ]
  ]  
    \end{forest}
    }
    \caption{Three types of SLAM methods.}
    \label{fig_tree_three_types}
\end{figure}
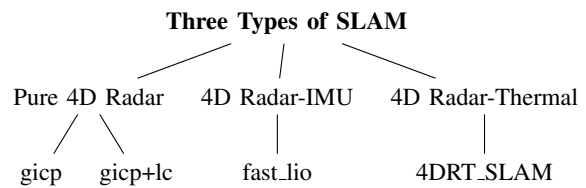

In this section, we will evaluate the performance of three types of SLAM algorithms (Fig.\ref{fig_tree_three_types}) with our dataset: 

\begin{enumerate}
    \item \textbf{Pure 4D radar}. We use our previous work \texttt{4DRadarSLAM}~\cite{Zhang2023ICRA}, using \texttt{gicp} for scan-to-scan matching. Meanwhile, loop closure (\texttt{lc}) can be added in to trigger graph optimization. So there are two options ``\texttt{gicp}" (w/o loop closure) and ``\texttt{gicp-lc}" (w/ loop closure).
    \item \textbf{4D radar - IMU fused}. We choose \texttt{Fast\_LIO}~\cite{Xu2021RAL}, which is originally proposed for LiDAR-IMU. We modified the input point cloud format to make it work with 4D radar. 
    \item  \textbf{4D radar - thermal camera}. We use our previous work \texttt{4DRT-SLAM}~\cite{Zhang2023CISRAM}. It follows the classical RGBD SLAM theory. Radar point cloud is projected onto the thermal image to get the depth image. Meanwhile, deep learnt features are extracted from the thermal images. Then, Perspective-n-Point (PnP) can be performed to calculate the odometry.
    
\end{enumerate}







\begin{table}[!h]
    \centering
    \caption{Quantitative analysis: trajectory error RE ($t_{rel}, r_{rel}$) and ATE ($t_{abs}$).}
    \label{tab_quantitative_error}
 \resizebox{0.49\textwidth}{!}{
    \begin{tabular}{l|lll|lll|lll|lll}
    \hline
       & \multicolumn{3}{c|}{gicp}    & \multicolumn{3}{c|}{gicp-lc}       & \multicolumn{3}{c|}{fast-lio}    & \multicolumn{3}{c}{4DRT-SLAM}      \\ \hline
     Dataset & \multicolumn{1}{c}{\begin{tabular}[c]{@{}c@{}}$t_{rel}$\\ (\%)\end{tabular}} & \multicolumn{1}{c}{\begin{tabular}[c]{@{}c@{}}$r_{rel}$\\ (deg/m)\end{tabular}} & \multicolumn{1}{c|}{\begin{tabular}[c]{@{}c@{}}$t_{abs}$\\ (m)\end{tabular}} & \multicolumn{1}{c}{\begin{tabular}[c]{@{}c@{}}$t_{rel}$\\ (\%)\end{tabular}} & \multicolumn{1}{c}{\begin{tabular}[c]{@{}c@{}}$r_{rel}$\\ (deg/m)\end{tabular}} & \multicolumn{1}{c|}{\begin{tabular}[c]{@{}c@{}}$t_{abs}$\\ (m)\end{tabular}} & \multicolumn{1}{c}{\begin{tabular}[c]{@{}c@{}}$t_{rel}$\\ (\%)\end{tabular}} & \multicolumn{1}{c}{\begin{tabular}[c]{@{}c@{}}$r_{rel}$\\ (deg/m)\end{tabular}} & \multicolumn{1}{c|}{\begin{tabular}[c]{@{}c@{}}$t_{abs}$\\ (m)\end{tabular}} & \multicolumn{1}{c}{\begin{tabular}[c]{@{}c@{}}$t_{rel}$\\ (\%)\end{tabular}} & \multicolumn{1}{c}{\begin{tabular}[c]{@{}c@{}}$r_{rel}$\\ (deg/m)\end{tabular}} & \multicolumn{1}{c}{\begin{tabular}[c]{@{}c@{}}$t_{abs}$\\ (m)\end{tabular}} \\ \hline
       cp    & 4.13 & 0.0552 & 3.96     & 2.79 & 0.0511 & \textbf{2.54} &    2.94 & 0.0468 & 2.67     & 13.22 & 0.1298 & {12.97}      \\
       garden  & 2.64 & 0.0310 & 4.53     & 2.38 & 0.0293 & \textbf{3.69}    & fail & fail & fail    & 5.58 & 0.0246 & {16.54}     \\
       nyl  & 4.62 & 0.0184 & 17.42   & 3.10 & 0.0120 & \textbf{14.34}    & 3.80 & 0.0208 & 21.10    & - & - & -     \\ \hline
       loop 1 & 12.99 & 0.0113 & 995.65    & 13.01 & 0.0113 & 996.03    & 12.26 & 0.0085 & \textbf{474.59}    & - & - & -      \\ 
       loop 2 & 4.84 & 0.0060 & 132.92     & 4.12 & 0.0065 & \textbf{68.88}    & 7.16 & 0.0057 & 159.00     & - & - & -     \\
       loop 3 & 3.22 & 0.0060 & \textbf{57.29}    & 3.51 & 0.0052 & 72.28    & 4.55 & 0.0064 & 77.53    & - & - & -      \\ 
      \hline
    \end{tabular}
    }
\end{table}


\begin{figure}[htbp]
	\centering
	\begin{subfigure}{.45\textwidth}
		\centering
    	\includegraphics[width=0.99\linewidth]{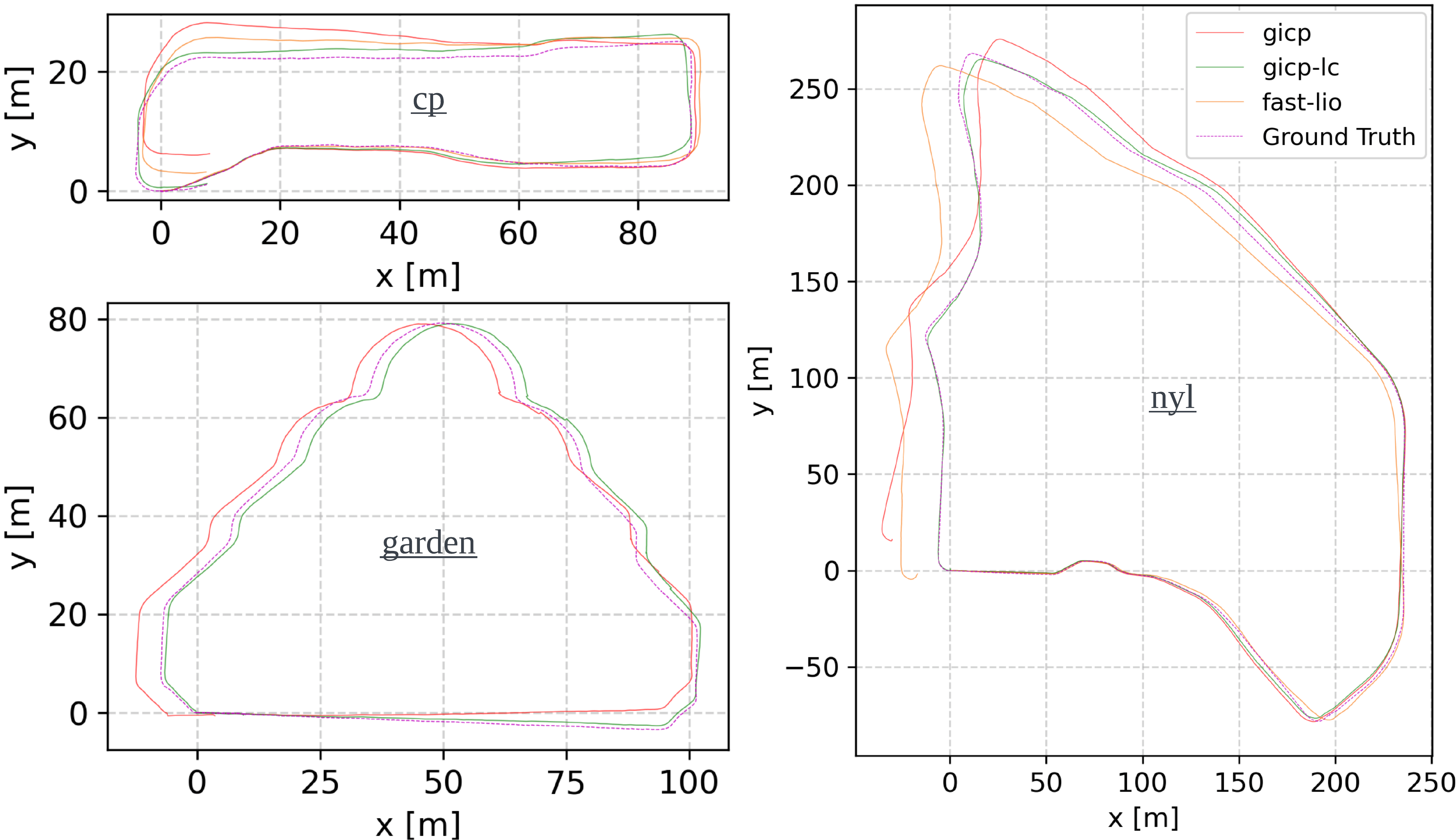}
    	\caption{On handcart (low speed), datasets “cp”,“garden”,“nyl”.}
    	\label{fig_qualitative_trajectory_handcart}
	\end{subfigure}  \\
	\begin{subfigure}{.45\textwidth}
		\centering
    	\includegraphics[width=0.99\linewidth]{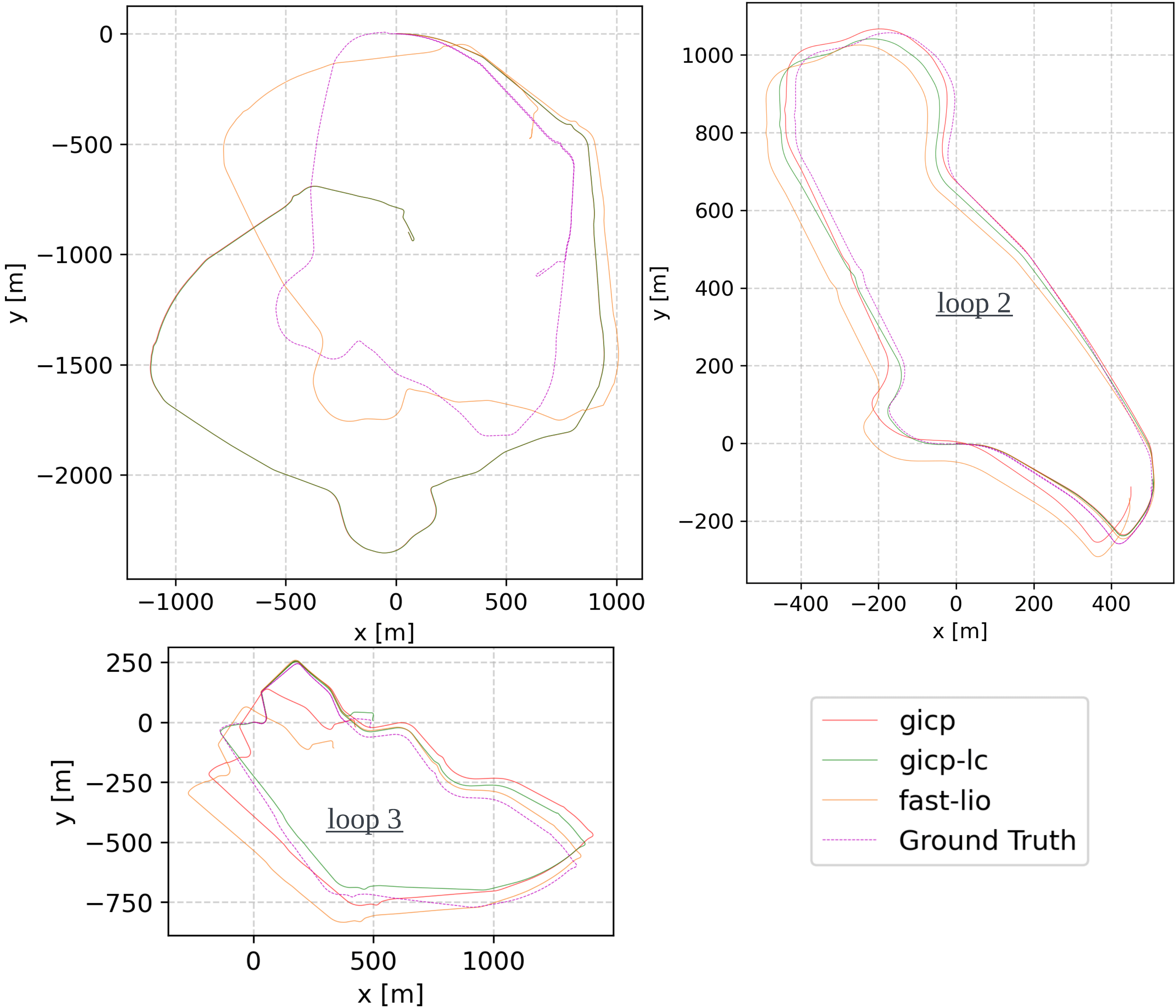}
    	\caption{On car (fast speed), datasets “loop 1”,“loop 2”,“loop 3”.}  
    	\label{fig_qualitative_trajectory_car}
	\end{subfigure}
	\caption{Compare the estimated odometry of ``gicp", ``gicp-lc" and ``fast-lio" with the ground truth, under the $6$ datasets.}
	\label{fig_qualitative_trajectory}
	\vspace{-8pt}
\end{figure}

\begin{figure}[htbp]
    \centering
	\begin{subfigure}{.2\textwidth}
		\centering
		\includegraphics[width=0.99\textwidth]{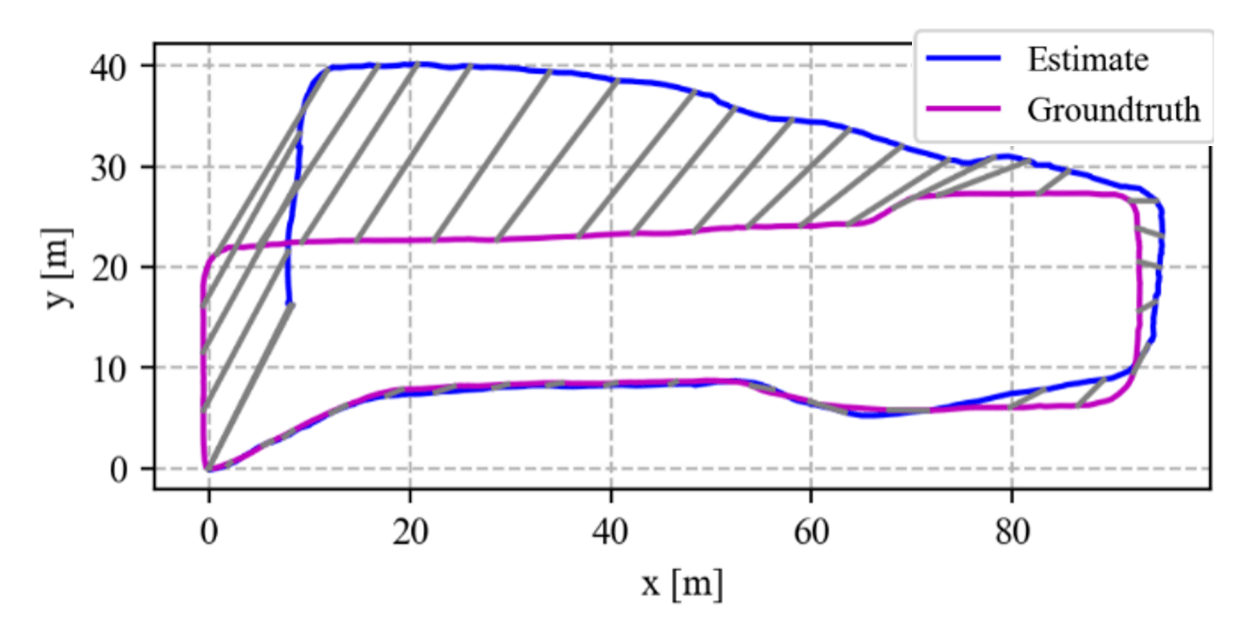}
		\caption{cp}
		\label{fig_quanti_CP}
	\end{subfigure} %
	\begin{subfigure}{.2\textwidth}
		\centering
		\includegraphics[width=0.99\textwidth]{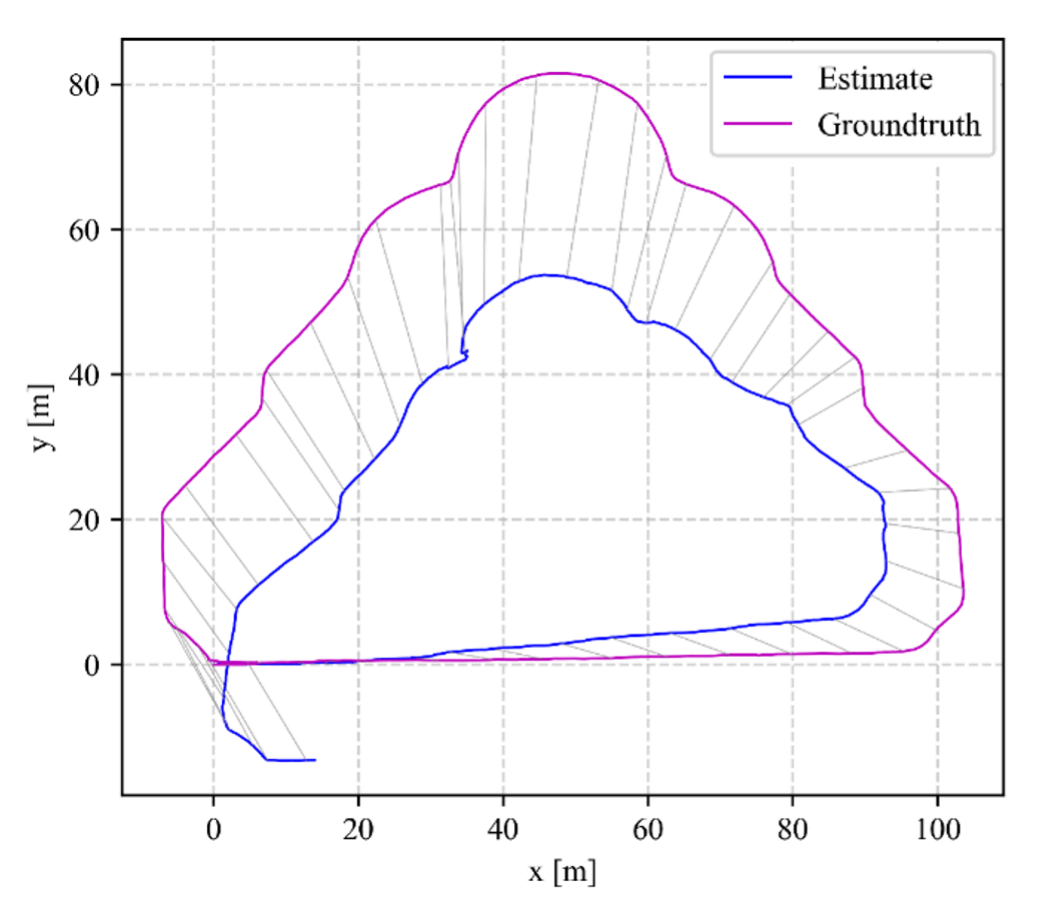}            
            \caption{garden}
		\label{fig_quanti_YNG}
	\end{subfigure}            
    \caption{Compare the  estimated odometry of ``4DRT-SLAM" vs. the  ground truth, under two datasets. }
    \label{fig_quanti_trajectory}
    \vspace{-12pt}
\end{figure}





\begin{figure*}[h]
	\centering
    \includegraphics[width=0.88\linewidth]{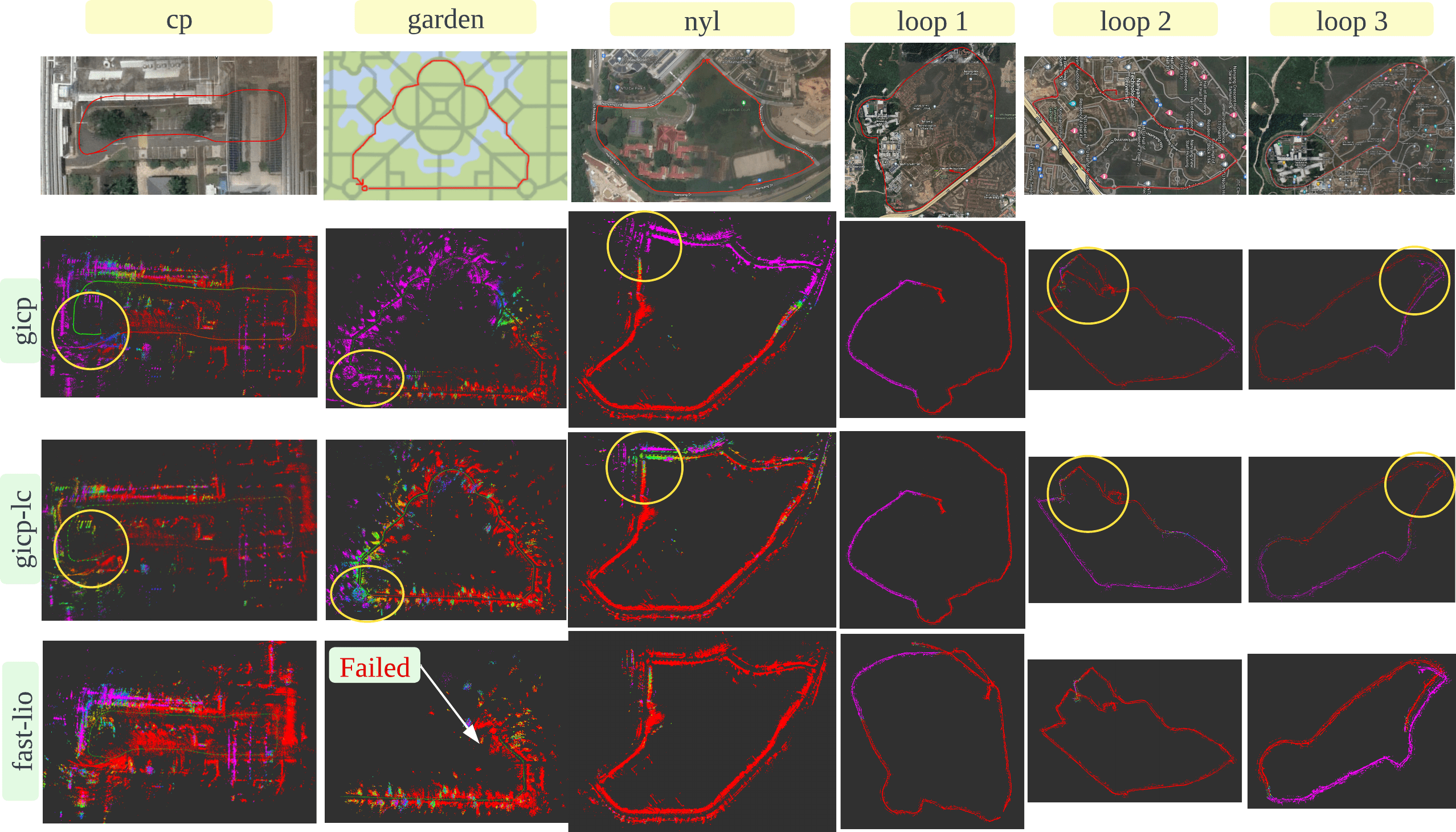}
	\caption{Visualization of the point cloud map of $6$ datasets. }  
	\label{fig_qualitative_map}
\end{figure*}

\subsection{Quantitative Analysis}
%


To evaluate trajectory error, we use the well-know \texttt{rpg\_trajectory\_evaluation}~\cite{Zhang18IROS_rpg} to compute both Absolute Trajectory Error (\textbf{ATE}) and Relative Error (\textbf{RE}). The quantitative results are shown in Tab.\ref{tab_quantitative_error}. 

For ``\texttt{gicp}", ``\texttt{gicp-lc}" and ``\texttt{fast-lio}", the experiments are performed for all $6$ routes, as shown in Fig.\ref{fig_qualitative_trajectory}. For ``\texttt{4DRT-SLAM}", since the performance is not that good, the experiments are only done with two small datasets ``cp" and ``garden", as shown in Fig.\ref{fig_quanti_trajectory}. {It can be observed that: }

\begin{enumerate}
    \item \textit{gicp performs better than fast-lio, except on dataset ``loop1". }  Through  analysis, possible explanation is:  gicp is a direct point cloud registration-based method, so it does not extract geometric features for odometry calculation. However, fast-lio is originally designed for LiDAR and it relies on plane and edge feature extraction for odometry calculation. Considering that 4D radar point cloud is much more noisy and sparse, it is more inaccurate to extract those features. Thus, it is not astonishing that fast-lio does not perform well on 4D radar.
    \item \textit{If loop closure is integrated, gicp-lc improves the performance significantly, compared with gicp.} This is straightforward, since valid loop closures are good constraints to optimize the global odometry. 
    \item \textit{fast-lio fails midway on the ``garden" dataset.} The possible reason for this is that fast-lio relies on plane feature extraction for odometry calculation. However, the garden is a very unstructured environment and  fewer plane features exist. Therefore, it may easily fail to extract valid plane features, thus it  may lose tracking and  fail midway.
    \item \textit{for dataset ``loop1", fast-lio performs best, the estimated trajectory of gicp-lc and gicp is the same.} This is because the loop closure is not triggered, thus graph optimization is not performed. Meanwhile, ``loop1" dataset is a semi-structured environment, thus, more valid plane features can be extracted so that fast-lio can work well. 
    \item \textit{4DRT-SLAM shows its effectiveness, but it performs the worst.}      This is mainly because 4DRT-SLAM is still in an early stage of development. There exists much space for improvement, for example, pre-processing the raw point cloud of 4D radar to reduce noisy points and avoid ghost points. 
\end{enumerate}

\subsection{Qualitative Analysis}
 For qualitative analysis, we visualize the point cloud maps built by ``\texttt{gicp}", 
``\texttt{gicp-lc}" and ``\texttt{fast-lio}" in Fig.~\ref{fig_qualitative_map}. 

\section{ Conclusions and Future Work } \label{sec_conclusions_and_future_work}
SLAM is entering a robust perception age, however, there are limited datasets that contain both 4D radar, thermal camera and IMU. It seriously hinders the research on robust SLAM in adverse conditions. Therefore, in this paper, we release the dataset \textit{NTU4DRadLM} to meet the urgent requirement. The dataset is collected in the campus of Nanyang Technological University, Singapore, for a total of around  {$17.6 km$, $85 mins$, $50 GB$}.  It includes all $6$ sensors: 4D Radar, thermal camera, IMU, 3D LiDAR, visual camera  and RTK GPS. The sensors are well-calibrated. It targets for both low-speed mobile robots and fast-speed unmanned vehicles. It covers structured, semi-structured and unstructured envrionments. The ground truth odometry is fine-tuned by fusing LiDAR SLAM, RTK GPS and loop closure. Three types of SLAM algorithms are evaluated on this dataset.  In future work, we will extend the dataset to include  adverse weather conditions. 

\addtolength{\textheight}{-3cm}   

\bibliographystyle{ieeetr}
\bibliography{ITSC2023_NTU4DRadLM}

\end{document}